\documentclass[journal]{IEEEtran}
\usepackage{amsmath}
\usepackage{amssymb}
\usepackage[ruled, linesnumbered]{algorithm2e}
\usepackage{subfigure}
\usepackage{graphicx}
\usepackage{url}
\usepackage{color}
\usepackage{cite}
\usepackage{diagbox}
\usepackage{booktabs}
\usepackage{multirow}
\graphicspath{{figures/}}
\usepackage{amsmath}
\usepackage{algorithmic}
\usepackage{eqparbox,array}
\makeatletter
\newcommand{\removelatexerror}{\let\@latex@error\@gobble}
\makeatother

\ifCLASSINFOpdf

\else
 
\fi

\pdfoutput=1

\begin{document}
	
\title{Causality-driven Sequence Segmentation for Enhancing Multiphase Industrial Process Data Analysis and Soft Sensing}

\author{Yimeng~He, Le~Yao,~\IEEEmembership{Member,~IEEE}, Xinmin Zhang,~\IEEEmembership{Member,~IEEE}, Xiangyin Kong and Zhihuan Song


\thanks{Yimeng He, Xinmin Zhang, Xiangyin Kong, and Zhihuan Song are with the State Key Laboratory of Industrial Control Technology, the College of Control Science and Engineering, Zhejiang University, Hangzhou 310027, China. (Email: yimenghe@zju.edu.cn, xinminzhang@zju.edu.cn, xiangyinkong@zju.edu.cn, songzhihuan@zju.edu.cn).

Le Yao is with the School of Mathematics, Hangzhou Normal University, Hangzhou 311121, China. (Email: yaole@hznu.edu.cn).
}
}

{}

\maketitle
\begin{abstract}
The dynamic characteristics of multiphase industrial processes present significant challenges in the field of industrial big data modeling. Traditional soft sensing models frequently neglect the process dynamics and have difficulty in capturing transient phenomena like phase transitions. To address this issue, this article introduces a causality-driven sequence segmentation (CDSS) model. This model first identifies the local dynamic properties of the causal relationships between variables, which are also referred to as causal mechanisms. It then segments the sequence into different phases based on the sudden shifts in causal mechanisms that occur during phase transitions. Additionally, a novel metric, similarity distance, is designed to evaluate the temporal consistency of causal mechanisms, which includes both causal similarity distance and stable similarity distance. The discovered causal relationships in each phase are represented as a temporal causal graph (TCG). Furthermore, a soft sensing model called temporal-causal graph convolutional network (TC-GCN) is trained for each phase, by using the time-extended data and the adjacency matrix of TCG. The numerical examples are utilized to validate the proposed CDSS model, and the segmentation results demonstrate that CDSS has excellent performance on segmenting both stable and unstable multiphase series. Especially, it has higher accuracy in separating non-stationary time series compared to other methods. The effectiveness of the proposed CDSS model and the TC-GCN model is also verified through a penicillin fermentation process. Experimental results indicate that the breakpoints discovered by CDSS align well with the reaction mechanisms and TC-GCN significantly has excellent predictive accuracy.
\end{abstract}


\IEEEpeerreviewmaketitle

\section{Introduction} \label{intro}
The multiphase process refers to a single process that encompasses multiple reaction phases, or is carried out under different conditions or in various devices \cite{wu2020self,jiang2019data}. This situation causes dynamic characteristics such as the relationships between variables to change with the transition of phases \cite{zhou2018multimode}. Traditional multivariate statistical models, such as principal component analysis (PCA) \cite{dong2018novel,dong2022denoising} and partial least squares \cite{zhao2006performance}, analyze the data of multiphase processes as a whole, overlooking the fact that the process characteristics vary significantly between different phases, which severely impacts the monitoring and predictive capabilities of the models \cite{zhang2015quality}. 

For a more profound comprehension of multiphase processes, numerous methods for phase segmentation have been introduced to separate the entire process into distinct phases. One way relies on expert knowledge, the process is divided into multiple segments based on different processing units and distinguishable operational stages within each unit \cite{dong1996batch}. This type of segmentation effectively reflects the operational states of processes. However, there are instances where the available prior knowledge is not sufficient to rationally divide the processes.

Therefore, data-driven segmentation methods are extensively applied to overcome the lack of process knowledge. The kernel change detection is an approach based on support vector machine, which detects the abrupt changes of signals in the feature space \cite{desobry2005online}. The Bayesian online change-point detection algorithm uses a modular approach with a discrete exponential prior over change-point intervals, enabling efficient inference of the current run length in sequential data \cite{alami2020restarted}. The mentioned models have harnessed machine learning techniques to achieve reliable segmentation of univariate time series, yet they have not been elegantly adapted for multivariable time series segmentation.

There are also numerous methods tailored for segmenting multivariate time series. Gaussian mixture model (GMM) is commonly used to divide the multiphase sequence into multiple phases \cite{liu2018sequential}, however, it simlply classifies the samples at each time stamp and ignores the time continuity during each phase. Camacho \textit{et al.} \cite{camacho2006multi} employed slice PCA to extract correlations between variables, and then clustered the segments of multivariate time series. Wang \textit{et al.} \cite{wang2019data} proposed a method that segments multivariate sequences according to the accuracy of predictive models. However, segmentation methods based on linear models may not be suitable for complex non-linear dynamic time series. Greedy Gaussian Segmentation (GGS) \cite{hallac2019greedy} searches over the possible breakpoints by reducing the covariance-regularized maximum likelihood problem to a combinatorial optimization problem. The accuracy and robustness of GGS surpass many other methods for segmenting multivariable time series. Nevertheless, these methods are sensitive to shifts in mean and variance, and may not be well-suited for non-stationary sequences exhibiting clear trends or periodicities. In summary, the models mentioned above all have certain limitations.

During a dynamical process, the mean, variance, and correlations of variables usually undergo significant changes, while the causal relationships between variables (also referred to as causal mechanisms) are much more stable, which reflect the underlying process information. Therefore, to ensure reliable  segmentation in most situations, causal mechanisms are used as a kind of stable features to assist in segmenting multivariate series. This article proposed a segmentation model, named as causality-driven sequence segmentation (CDSS) model. The CDSS model integrates causal discovery \cite{pearl2009causal, assaad2022survey, sun2023nts} with the construction of predictive models. This model separates the sequence into distinct phases by identifying shifts in the causal interactions between various variables. Such an approach maintains continuity in the time dimension and simultaneously improves the interpretability of segmentation results. Furthermore, the temporal causal graphs, which are extracted during the segmentation, can be used to assist the development of quality predictive models.

Graph-based deep learning models, like graph convolution (GC) network \cite{guo2021hierarchical}, are widely applied for time series forecasting in the field of industrial soft sensing \cite{jia2023graph}. Chen \textit{et al.} \cite{chen2024residual} proposed an industrial soft sensing model which leverages self-attention mechanisms for graph mining and employs GC layers to integrate variable relationships. Wang \textit{et al.} \cite{wang2023interpretable} proposed a novel stacked graph convolutional network (S-GCN) to explore the potential correlations between process variables and enhance the model interpretability. Causal discovery offers sufficient and robust insights into the relationships between variables, enabling the construction of causal graphs \cite{zheng2023spatiotemporal}. It facilitates the industrial application of causal graphs in areas such as soft sensing \cite{yu2022stable}. For example, causal graphs assist to select measuring variables related to quality variables for soft sensing modeling \cite{he2022neural}, which improve the accuracy of predictions. Therefore, in order to improve the performance and interpretability of soft sensing models, the proposed TC-GCN model integrates the temporal and spatial features by GC layers.

The contribution of this article can be summarized as:

(1) The proposed CDSS model splits sequences when causal mechanisms abruptly change. It effectively divides the sequence into different operational phases.

(2) The metric, similarity distance, is designed to detect the consistency of causal mechanisms over time, which includes the causal similarity distance and the stable similarity distance.

(3) The proposed TC-GCN soft sensing model integrates both the temporal and spatial features by transferring the sequence data and the adjacency matrix of the TCG to the GC layer, enhancing the predictive accuracy.

The rest of this article is organized as follows. In Section \ref{sec:preliminary}, we give a preliminary about temporal causal discovery. In Section \ref{sec:method1}, the proposed CDSS model and the similarity distance are presented. In Section \ref{sec:method2}, the proposed TC-GCN model is described. In Section \ref{case study}, we use numerical examples and the penicillin fed-batch fermentation process to test the performance of CDSS and TC-GCN. Finally, the conclusions are written in Section \ref{sec:conclusions}.

\section{Preliminary of Temporal Causal Discovery}\label{sec:preliminary}

This section introduces an original temporal causal discovery model, NTS-NOTEARS \cite{sun2023nts}. It is used for the development of the proposed CDSS model. The NTS-NOTEARS-based model consists of 1D-convolutional neural networks (CNNs), which is designed to discover the instantaneous and lagged variable dependencies for multivariable time series. 
Each CNN is comprised of the 1D-convoluntinal layer and the fully-connected layers. Suppose that the time series includes $d$ variables, there are $d$ CNNs jointly trained. For each time step $t \geq K+1$, the $j$-th CNN predicts the expectation of the target variable $x^t_j$, given all preceding variables up to the time step $t-K$ (denoted as $\boldsymbol{x}^{t-k}$) and all variables at the same time step $t$ other than $x^t_j$ (denoted as $\boldsymbol{x}^t_{-j}$):
\begin{equation}
	E \left[x^t_j | PA(x^t_j)\right] = \textit{CNN}_j({\boldsymbol{x}^{t-k}: 1\leq k \leq K}, \boldsymbol{x}^t_{-j})
	\label{eq:cnn}	
\end{equation}
where $PA(x^t_j)$ denotes the parents of $x^t_j$ that are determined by the CNN weights (see next paragraph). Here $K$ is the hyperparameter denoting the maximum time lag. The convolution layer of the CNN consists of $m$ kernels, stride equal to 1 and no padding. The shape of each convolutional kernel is $d \times (K+1)$ where the last column $K+1$ represents instantaneous connections.

The kernel weights of the CNNs are transformed to the elements of the weighted adjacency matrix $\boldsymbol{W}$. Each entry $W^k_{ij}$ in $\boldsymbol{W}$ represents the dependency strength of the directed edge from variable $x^k_i$ to variable $x^{K+1}_j$, which is formulated as the $L^2$-norm of the kernel weights:
\begin{equation}
	W^k_{i,j} = \|\boldsymbol{\phi}^k_{i,j}\|_2 \hspace{.2cm} \textit{for} \hspace{.2cm} k=1, \cdots, K+1
\end{equation}
where $\boldsymbol{\phi}^k_{i,j}$ represents $m$ kernel weights corresponding to the input $x^k_i$ in the $j$-th CNN. Finally, weight threshold $w^k_{thres}$ is applied to each time step $k$ to prune with weak dependency strengths and define the parent set of each variable.

The training objective comprises four parts: 1) the predictive loss of the child values given the parents; 2) a sparsity penalty for the CNN weights; 3) a regularization term for all parameters; 4) a cyclicity penalty to driven the induced weights to define an acyclic graph. It is formulated as follows:

\begin{small}
\begin{align}
	\nonumber
    &F(\boldsymbol{\Theta})=L_{train} + \sum_{k=1}^{K+1} \lambda^k_1 \cdot \| \boldsymbol{\phi}^k_j \|_{1} + \frac{1}{2} \lambda_2 \cdot \| \boldsymbol{\theta}_j \|_2 + h(\boldsymbol{W}^{K+1}) \\
    \nonumber
    &L_{train}= \frac{1}{T-K}\cdot\sum_{t=K+1}^{T} \sum_{j=1}^{d}  \mathcal{L}(x^{t}_j, \textit{CNN}_j({\boldsymbol{x}^{t-k}: 1\leq k \leq K}, \boldsymbol{x}^t_{-j}))           \\
    &h(\boldsymbol{W}^{K+1})= tr(e^{\boldsymbol{W}^{K+1} \odot \boldsymbol{W}^{K+1}}) - d
\end{align}
\end{small}where $\mathcal{L}()$ denotes the least-squares loss, $\boldsymbol{\phi}^k_j$ is the concatenation of the $\boldsymbol{\phi}^k_{i,j}$ vector, and $\boldsymbol{\theta}_j$ denotes all parameters of the $j$-th CNN. The $tr(A)$ and $e^A$ are the trace and matrix exponential of matrix A, respectively, and $\odot$ is element-wise product. The function $h$ enforces the acyclicity constraint between instantaneous dependencies.  
\section{Proposed Causality-driven Sequence Segmentation Model} \label{sec:method1}
\begin{figure}[t]
	\centering   
	\includegraphics[width=\columnwidth]{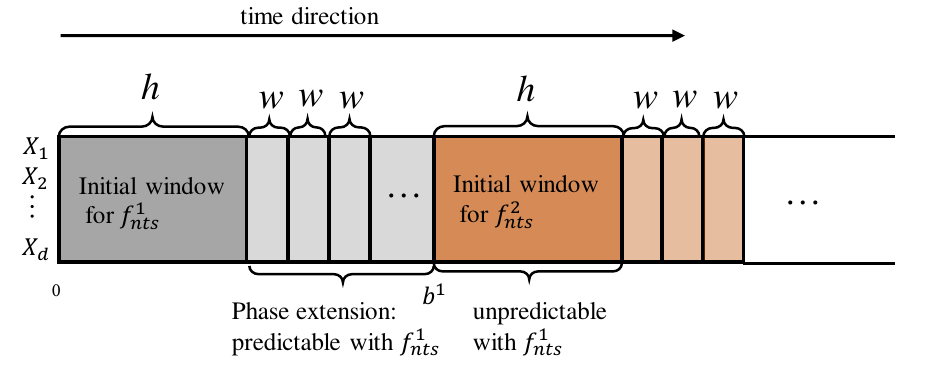} 
	\caption{Illustration of phase segmentation.} \label{fig:phase-segmentation}
\end{figure}
The proposed causality-driven sequence segmentation (CDSS) model separates the sequence by identifying the breakpoints where the previous causal mechanisms no longer fit the latest testing samples. As illustrated in Fig. \ref{fig:phase-segmentation}, the CDSS model first trains a NTS-NOTEARS-based predictive model \cite{sun2023nts}, using the samples from an initial window at the start of each phase. Then, the predictive model gradually tests the next $w$ samples. A transition to a new phase is indicated when the latest testing samples are not predictable by the established predictive model.
\begin{algorithm}
	\renewcommand{\algorithmicrequire}{\textbf{Input:}}
	\renewcommand{\algorithmicensure}{\textbf{Output:}}
	\renewcommand\algorithmiccomment[1]{%
	\hfill\#\ \eqparbox{COMMENT}{#1}%
		} 
	\newcommand\LONGCOMMENT[1]{%
	\hfill\#\ \begin{minipage}[t]{\eqboxwidth{COMMENT}}#1\strut\end{minipage}%
		} 
	\caption{The procedure of CDSS }	
	\begin{algorithmic}[1]
		\REQUIRE{a sequence $\boldsymbol{X}$=[$\boldsymbol{x}_0$, $\boldsymbol{x}_1$, \dots, $\boldsymbol{x}_T$]; 
			
			hyper parameters : $h$, $K$, $w$, $\zeta$, $\alpha$, $\beta$, $N_{max}$, $L_{min}$.}
		\STATE Initialize a breakpoint list, \textit{bpt} = [0,];
		\STATE Initialize a causal graph list, \textit{cg} = [None,];
		\WHILE {\textit{len}(\textit{bpt}) < $N_{max}$} 
		\STATE \textit{bpt},\textit{cg} $\gets$ AddNewBreakpoint($\boldsymbol{X}$, \textit{bpt}, \textit{cg}, $h$, $K$, $w$, $\zeta$, $\alpha$, $\beta$);\\
		\COMMENT{\textit{Repetition of Algorithm2}}
		\IF{\textit{bpt}[-1] + $L_{min}$ > T}
		\STATE \textit{break};
		\ENDIF
		\ENDWHILE
		\RETURN \textit{bpt}, \textit{cg}.
	\end{algorithmic}
\end{algorithm}

\begin{algorithm}
	\renewcommand{\algorithmicrequire}{\textbf{Input:}}
	\renewcommand{\algorithmicensure}{\textbf{Output:}}
	\renewcommand\algorithmiccomment[1]{%
	\hfill\#\ \eqparbox{COMMENT}{#1}%
		} 
	\newcommand\LONGCOMMENT[1]{%
	\hfill\#\ \begin{minipage}[t]{\eqboxwidth{COMMENT}}#1\strut\end{minipage}%
		} 
	\caption{AddNewBreakpoint}
	\begin{algorithmic}[1]
		\REQUIRE{a sequence $\boldsymbol{X}$=[$\boldsymbol{x}_0$, $\boldsymbol{x}_1$, \dots, $\boldsymbol{x}_T$]; 
			
			a breakpoint list, \textit{bpt}=[0, $b^1$, \dots, $b^{p-1}$];
			
			a causal graph list, \textit{cg}=[None, $G^1$, \dots, $G^{p-1}$];
			
			hyper parameters : $h$, $K$, $w$, $\zeta$, $\alpha$, $\beta$.}
		\IF{$b^{p-1} +  h + w> T$}
		\STATE $b^{p-1} \gets T$
		\RETURN
		\ENDIF
		\STATE Initialize a time window  $\boldsymbol{X}^p=\{ \boldsymbol{x}^p_t \}$, $t=1,2,\cdots, h$, where $\boldsymbol{x}^p_t = \boldsymbol{x}_{b^{p-1}+t}$;
		\STATE Get the normalized $\boldsymbol{\bar{X}}^p$ and the mean $\boldsymbol{\mu}^p_{train}$;\\
		\COMMENT{\textit{Initialization}}
		\STATE Initialize the NTS-NOTEARS-based predictive model $f_{nts}^p$ with the maximum time lag $K$;
		\STATE Train $f_{nts}^p$ with $\boldsymbol{\bar{X}}^p$, get the temporal causal graph $G^{p}$;\\
		\COMMENT{\textit{Model Training}}
		\STATE Calculate the training loss $L^p_{train}$;
		\STATE Calculate the threshold for the $p$-th phase, $\rho^p$;
		\STATE $n=1$;
		\WHILE {$b^{p-1}+h+n \cdot w<T$}
		\STATE Get the current time window $\boldsymbol{X}^{p-test} = \{\boldsymbol{x}_t^{p-test}\}$, $\boldsymbol{x}_t^{p-test}=\boldsymbol{x}_{b^{p-1} + h+t}$, $t=1,2,\cdots, n \cdot w$;
		\STATE Get the normalized $\bar{\boldsymbol{X}}^{p-test}$ and the mean $\boldsymbol{\mu}^p_{test}$;
		\STATE Calculate the causal similarity distance $Dist_c^p$;
		\STATE Calculate the stable similarity distance $Dist_m^p$;
		\STATE Calculate the similarity distance $Dist^p$;  
		\IF{$Dist^p < \rho^p$}
		\STATE $n \gets n+1$;
		\COMMENT{\textit{Phase Extension}}
		\ELSE
		\STATE \textit{break};
		\ENDIF
		\ENDWHILE
		\STATE $b^p \gets b^{p-1} + h + n \cdot w$;
		\STATE $\textit{bpt} \gets \textit{bpt} \cup b^p$;
		\STATE $\textit{cg} \gets \textit{cg} \cup G^p$;
		\RETURN \textit{bpt}, \textit{cg}.	
	\end{algorithmic}
\end{algorithm}
In general, the procedure of CDSS is represented in Algorithm 1 and the detailed procedure of adding a new breakpoint is displayed in Algorithm 2. Specifically, there is a time series $\boldsymbol{X}=[\boldsymbol{x}_0, \boldsymbol{x}_1, \cdots, \boldsymbol{x}_T]$, the procedure of CDSS for phase segmentationn is described as follows:

\textbf{\textit{Initialization.}} The samples in a time window of length $h$ are used to discover the causal mechanisms of the $p$-th phase. The data set is denoted as $\boldsymbol{X}^p=\{ \boldsymbol{x}^p_t \}$, $t=1,2,\cdots, h$, where $\boldsymbol{x}^p_t = \boldsymbol{x}_{b^{p-1}+t}$, and $b^{p-1}$ is the breakpoint between the ($p$-1)-th phase and the $p$-th phase, as well as the starting point of the $p$-th phase. Before training the predictive model, $\boldsymbol{X}^p$ is normalized to have the zero mean and unit standard deviation, denoted as $\boldsymbol{\bar{X}}^p$. The mean vector of $\boldsymbol{X}^p$ is denoted as $\boldsymbol{\mu}^p_{train}$.

\textbf{\textit{Model Training.}} The NTS-NOTEARS-based predictve model $f_{nts}^p$ is trained using the normalized window data $\bar{\boldsymbol{X}}^p$. Through the training process, the temporal causal graph of the $p$-th phase $G^p$ is obtained. The training loss $L^p_{train}$ is formulated as follows:
\begin{equation}
	L^p_{train} = \sqrt{\frac{1}{h-K}\sum_{t=K+1}^{h}\|\hat{\boldsymbol{x}}^p_t - \bar{\boldsymbol{x}}^p_t \|_2}
\end{equation}
where $\hat{\boldsymbol{x}}^p_t \in R^d$ is the predictive vector of $\bar{\boldsymbol{x}}^p_t$, and each element of $\hat{\boldsymbol{x}}^p_t$ is the predictive value obtained from the CNN mentioned in Eq. (\ref{eq:cnn}).

\textbf{\textit{Phase Extension.}} The CDSS expands the duration of the phase by $w$ time steps, continuing until it encounters a breakpoint where the CNNs of the NTS-NOTEARS-based model no longer accurately forecast the latest samples. This breakpoint signifies a transition where the existing causal mechanisms fail to align with the current data, a new phase is supposed to start. A metric, named as similarity distance, is defined to determine whether the present samples correspond to the previous phase. The similarity distance encompasses two parts, the causal similarity distance and the stable similarity distance. 

The causal similarity distance quantifies the difference in causal relationships between the initial data window and the current data window, expressed as the testing loss for the latest samples:
\begin{equation}
	Dist_c^p = \sqrt{\frac{1}{n \cdot w}\sum_{t=1}^{n \cdot w}\|\hat{\boldsymbol{x}}^{p-test}_t - \bar{\boldsymbol{x}}^{p-test}_t \|_2}
\end{equation}
where $\boldsymbol{X}^{p-test} = \{\boldsymbol{x}_t^{p-test}\}$, $\boldsymbol{x}_t^{p-test} = \boldsymbol{x}_{b^{p-1}+h+t}$, $t=1,2,\cdots, n \cdot w$ denotes the current samples, and $n$ is the moving times of the testing data window. Here, $\bar{\boldsymbol{x}}^{p-test}_t$ means the normalized $\boldsymbol{x}_t^{p-test}$, and $\hat{\boldsymbol{x}}^{p-test}_t$ denotes the predictive value of $\bar{\boldsymbol{x}}^{p-test}_t$ from the model $f_{nts}^p$. 

The stable similarity distance is used to evaluate the distance of stable states between the initial window and the current window, which is represented as	 the Manhattan distance between the mean vectors of the two data windows:
\begin{equation}
	Dist_m^p = \|\boldsymbol{\mu}^p_{train} - \boldsymbol{\mu}^p_{test} \|_1
\end{equation}
where $\boldsymbol{\mu}^p_{train}$ denotes the mean vector of $\boldsymbol{X}^p$, and $\boldsymbol{\mu}^p_{test}$ denotes the mean vector of $\boldsymbol{X}^{p-test}$.

To comprehensively considers $Dist_c^p$ and $Dist_m^p$, the coefficient $\zeta$ is set to balance the two items (generally, scaling the two distances to a similar range is recommended). Therefore, the similarity distance is formulated as follows:
\begin{equation}
	Dist^p = Dist_c^p + Dist_m^p/ \zeta
\end{equation}
when $Dist^p$ is up to the threshold $\rho^p$, the time stamp is regarded as the breakpoint between the $p$-th phase and the ($p$+1)-th phase, denoted as $b^p$. The $\rho^p$ is defined as follows:
\begin{equation}
	\rho^p = \alpha \cdot L^p_{train} + \beta
\end{equation}
where $\alpha \cdot L^p_{train}$ indicates the limit of $Dist_c^p$, while $\beta$ indicates the limit of $Dist_m^p$.

\textbf{\textit{Repetition.}} The steps of \textit{initialization}, \textit{model training} and \textit{phase extension} are supposed to be repeated until the number of breakpoints reaches the maximum $N_{max}$ or the length of the remaining sequence is less than the minimum $L_{min}$.

\section{Proposed Temporal-causal Graph Convolutional Network for Soft Sensing} \label{sec:method2}
\begin{figure}[t]
	\centering   
	\includegraphics[width=0.7\columnwidth]{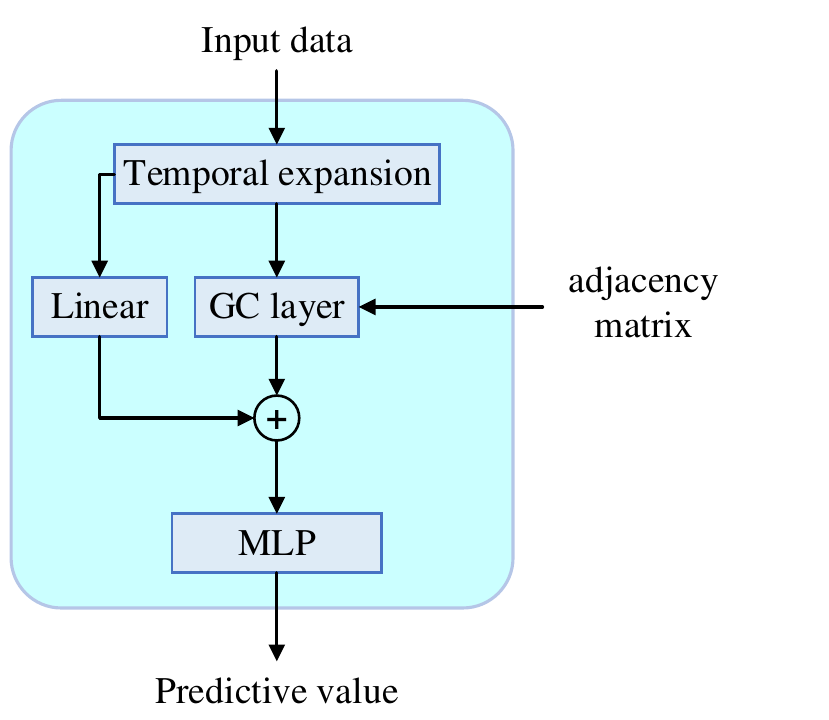} 
	\caption{The structure of TC-GCN.} \label{fig:tc-gcn}
\end{figure}

The proposed CDSS model partitions the time series into different phases, and the temporal causal graph for each phase is obtained. For the purpose of quality prediction, a novel soft sensing model called temporal-causal graph convolutional network (TC-GCN) has been developed for each phase. The TC-GCN integrates both the temporal and spatial features, by passing the sequence data and the adjacency matrix of the temporal causal graph to the graph convolutional (GC) layer \cite{guo2021hierarchical}. The structure of TC-GCN is shown in Fig. \ref{fig:tc-gcn}.

In order to match the dimension of the temporal causal graph, the phase data $\boldsymbol{X}^p$ is extended to $\tilde{\boldsymbol{X}}^p$ $\in$ $R^{N \times ((K+1) \times d)}$, each sample $\tilde{\boldsymbol{x}}^p_t$ consists of $\boldsymbol{x}^p_{t-K},\boldsymbol{x}^p_{t-K+1}$, $\cdots$,  $\boldsymbol{x}^p_{t}$, and $N$ is the number of samples in this phase. 

The GC layer can be written as:
\begin{equation}
	\boldsymbol{h}^{GC} = \boldsymbol{D}^{-0.5}\boldsymbol{A}_p^{-0.5}\boldsymbol{D}^{-0.5}\tilde{\boldsymbol{X}}^p\boldsymbol{W}_{enc}^{GC}
\end{equation}
where $\boldsymbol{A}_p$ is the adjacency matrix of the temporal causal graph $G^{p}$, $\boldsymbol{W}_{enc}^{GC}$ is the encoder parameter matrix which is obtained through the training, and $\boldsymbol{D}$ is the degree matrix of the graph, given as follows:
\begin{equation}
	\boldsymbol{D}_{[i,i]}=\sum_j{\boldsymbol{A}_{p[i,j]}}
\end{equation}

Unlike the conventional GCN, a residual structure is used after the GC layer. The residual information is passed through a simple linear projection layer, and then added to the output of the GC layer. The input of the following multiple perceptron (MLP) is denoted as $\boldsymbol{h}_{0}$, given as:
\begin{equation}
	\boldsymbol{h}_{0} =\boldsymbol{h}^{GC} + \tilde{\boldsymbol{X}}^p \boldsymbol{W}_{res}^{GC}
\end{equation}
where $\boldsymbol{W}_{res}^{GC}$ is the trainable weight matrix of the linear projection layer. The $l$-th layer of MLP is formulated as:
\begin{equation}
	\boldsymbol{h}_{l} =ReLU(\boldsymbol{h}_{l-1} \boldsymbol{W}_{l}^T + \boldsymbol{b}_{l})
\end{equation}
where $\boldsymbol{h}_{l-1}$ is the output of the ($l$-1)-th layer. Here, $\boldsymbol{W}_{l}$ and $\boldsymbol{b}_{l}$ are the trainable weight matrix and bias respectively.

The output of the last layer of MLP is the predictive value $\hat{y}$, which also represents the estimated quality variable. The mean square error (MSE) is used as the loss function, which is presented as follows:
\begin{equation}
	Loss = \frac{1}{N} \sum_{i=1}^{N} |y_i-\hat{y}_i|^2
\end{equation}
\begin{figure}[t]
	\centering   
	\includegraphics[width=0.9\columnwidth]{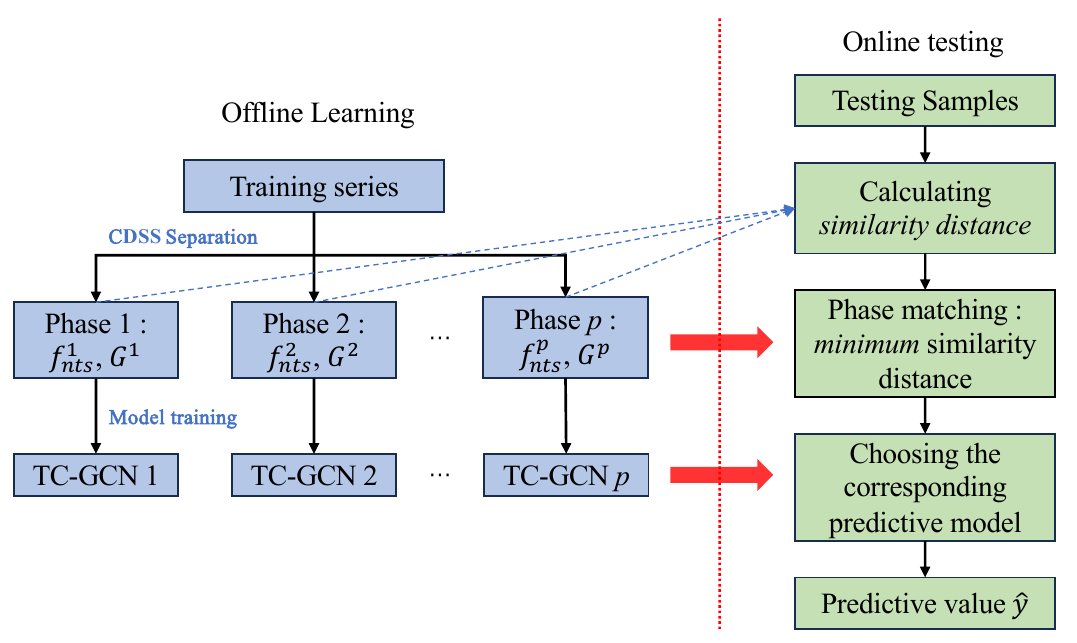} 
	\caption{The workflow of soft sensing.} \label{fig:soft_sensing procedure}
\end{figure}

The comprehensive soft sensing workflow is illustrated in Fig. \ref{fig:soft_sensing procedure}, including two steps: offline training and online testing. During the offline training, the training sequence is segmented into different phases using the CDSS model. Subsequently, phase-specific data and temporal causal graphs (TCGs) are derived through the training of the NTS-NOTEARS-based predictive models. In the online testing step, each test sample is matched with the most fitting phase by calculating its similarity distances to all the existing phases. After that, the corresponding TC-GCN model is applied to predict the quality variable of the testing sample. The procedure is summarized as segmenting-matching-predicting.

\section{Case Studies}\label{case study}
\subsection{Numerical Example} 
This section utilizes numerical examples to validate the proposed CDSS method. Here, a stationary example and a non-stationary example are generated respectively.
\subsubsection{The Stationary Numerical Example}
A stationary multivariable sequence consisting of three continuous modes is generated. There are totally 5 variables (nodes) in this case. The generation process of each node can be formulated as the following equation:
\begin{align}
	\nonumber
	x_i^t =& tanh(PA(x_i^t) \cdot \boldsymbol{W}^1_i) + cos(PA(x_i^t) \cdot \boldsymbol{W}^2_i) \\
	&+ sin(PA(x_i^t) \cdot \boldsymbol{W}^3_i) + z_i,i=1,\cdots,d
\end{align}
where $x_i^t$ denotes the value of the node $i$ at time stamp $t$, $PA(x_i^t)$ are the parent nodes of $x_i^t$, which are identified by a pre-defined TCG. $\boldsymbol{W}^1_i, \boldsymbol{W}^2_i, \boldsymbol{W}^3_i$ are randomly pre-defined weight vectors, varying for different nodes, but do not change over time during a certain phase. $z_i$ is the \textit{iid} Gaussian noise, $z_i \sim N(0,1)$.

\begin{figure}[t]
	\centering 
	\subfigure[TCG of mode 1.]{  
		\includegraphics[width=0.25\columnwidth]{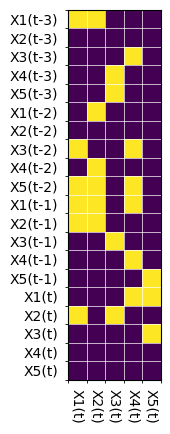} \label{fig:True_tcn1} }
	\subfigure[TCG of mode 2.]{  
		\includegraphics[width=0.25\columnwidth]{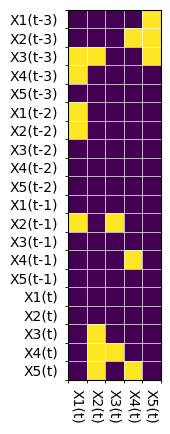} \label{fig:True_tcn2}}
	\subfigure[TCG of mode 3.]{  
		\includegraphics[width=0.25\columnwidth]{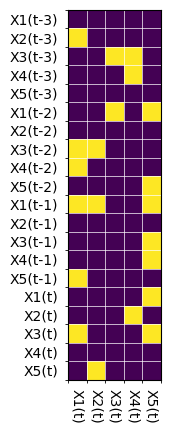} \label{fig:True_tcn3.png}}
	
	\caption{The pre-defined TCGs of the three modes.} \label{fig:true tcn}
\end{figure}

First, the pre-defined TCGs are randomly generated for the three modes respectively, which are displayed in Fig. \ref{fig:true tcn}. As Fig. \ref{fig:true tcn} shows, the maxinum time lag is set as 3, and it displays whether $x_{i}^{t}$ depends on $x_{i}^{t-k}$ and $x_{i-}^{t}$ (the subscript $i-$ means all the other nodes excpet node $i$). The 500 samples from each pattern are selected and then concatenated in order to form a multi-modal time series consisting of 1500 samples. The indexes of the two breakpoints are 500 and 1000.

It can also be seen that the means and variances of these three segments are very similar, and the main difference between the three modes is the dependencies between variables (reflected in Fig. \ref{fig:true tcn}) and the weight vectors for all nodes ($\boldsymbol{W}^1_i, \boldsymbol{W}^2_i, \boldsymbol{W}^3_i$). Therefore, experts and some traditional classification methods cannot clearly identify the breakpoints. For example, Gaussian mixture model (GMM) cannot segment the sequence into three consecutive segments \cite{shao2019semisupervised}.

\begin{figure}[t]
	\centering   
	\includegraphics[width=\columnwidth]{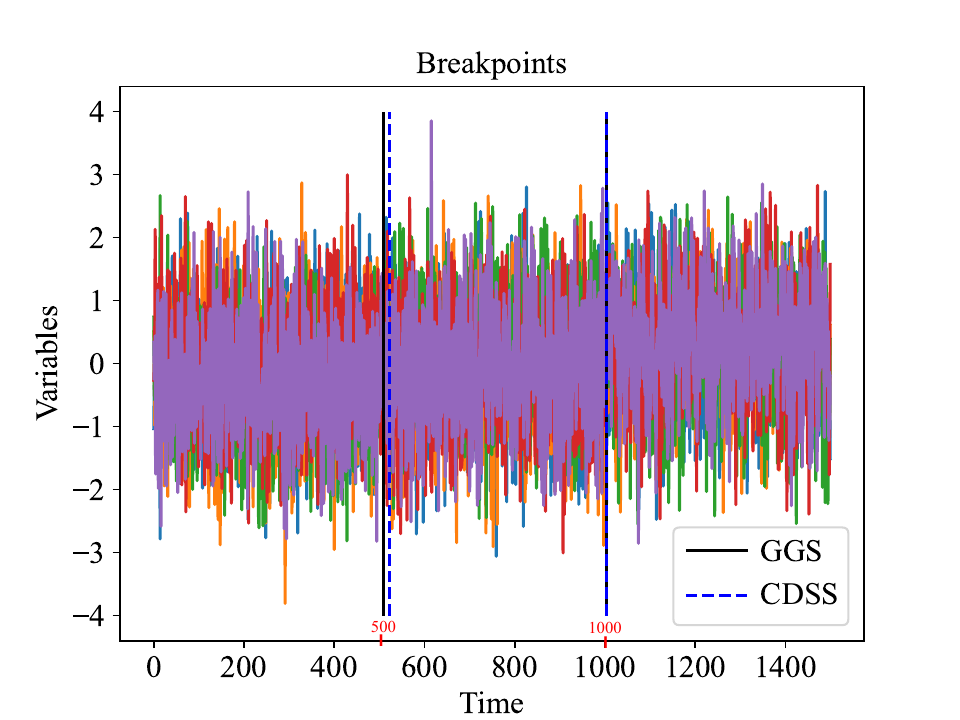} 
	\caption{The segmentation results for stationary numerical example.} \label{fig:num_stable}
\end{figure}

Fig. \ref{fig:num_stable} shows that the breakpoints discovered by Greedy Gaussian segmentation (GGS) \cite{hallac2019greedy} and the CDSS model. The GGS is a powerful approach for multivariable time series segmentation. The breakpoints found by GGS are 509 and 1004, marked with black solid line. The breakpoints discovered by CDSS are 522 and 1003, marked with blue dashed line. The segmentation results of the two methods are both very close to the ground truth, 500 and 1000, which indicates that GGS and CDSS both have excellent performance on stationary multivariable time series segmentation.  

\subsubsection{The Non-stationary Numerical Example}
Unstable time series are also commonly encountered in practical scenarios such as industry and finance. Therefore, a non-stationary numerical example is designed to further verify the effectiveness of the proposed CDSS model. To simplify the design complexity, the lagged dependencies between variables are not considered. This numerical example involves three variables. The dependencies between the nodes in the three modes are described as follows respectively:

\textbf{Mode 1}: $w_1=0.05, t=1\sim500, z_1,z_2,z_3\sim N(0.1,0.05)$
\begin{align}
	\nonumber
	y_1(t) =& sin(w_1t) + z_1 \\
	\nonumber
	y_2(t) =& 1.2y_1(t) + z_2 \\
	y_3(t) =& 0.5y_1(t)^2 + z_3
\end{align}

\textbf{Mode 2}: $w_2=0.05, t=1\sim500, z_1,z_2,z_3\sim N(0.1,0.05)$
\begin{align}
	\nonumber
	y_1(t) =& sin(w_2t) + z_1 \\
	\nonumber
	y_2(t) =& 0.6y_1(t) + z_2 \\
	y_3(t) =& 0.5y_1(t)^2 + 0.6y_2(t) + z_3
\end{align}

\textbf{Mode 3}: $w_3=0.03, t=1\sim500, z_1,z_2,z_3\sim N(0.1,0.05)$
\begin{align}
	\nonumber
	y_1(t) =& sin(w_3t) + z_1 \\
	\nonumber
	y_2(t) =& 0.6y_1(t) + z_2 \\
	y_3(t) =& 0.5y_1(t)^2 + 1.0y_2(t) + z_3
\end{align}

\begin{figure}[t]
	\centering   
	\includegraphics[width=\columnwidth]{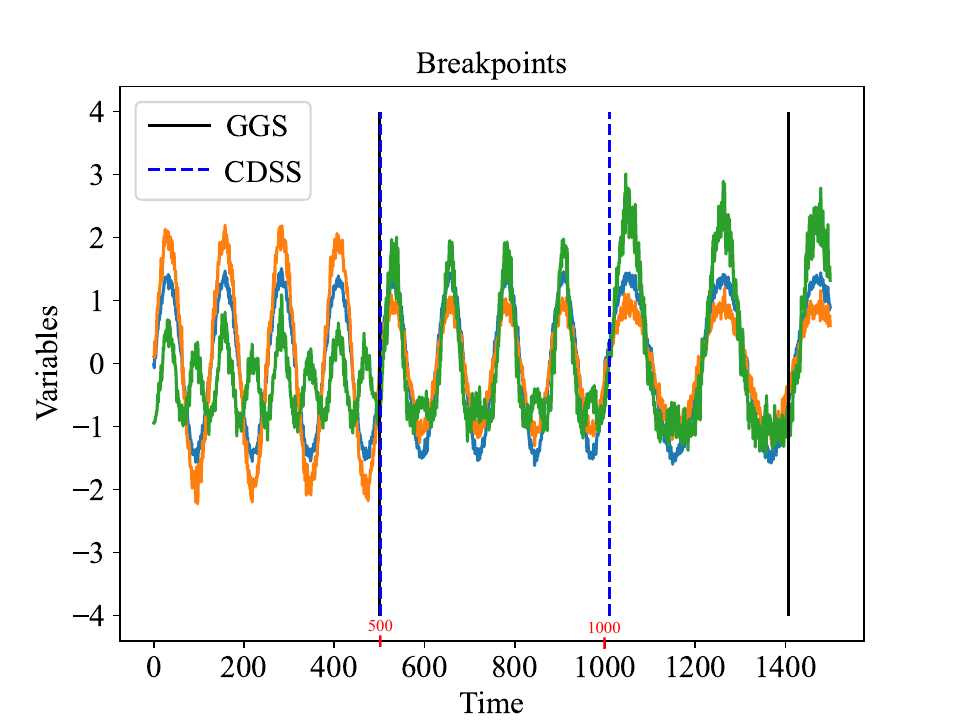} 
	\caption{The segmentation results for non-stationary numerical example.} \label{fig:num_unstable}
\end{figure}

Fig. \ref{fig:num_unstable} displays that the breakpoints discovered by GGS and CDSS. The breakpoints found by GGS are 501 and 1406, marked with black solid line. The breakpoints discovered by CDSS are 504 and 1010, marked with blue dashed line. The second breakpoint identified by GGS deviates significantly from the true breakpoint. It indicates that GGS is unable to capture the transition between Mode 2 and Mode 3. However, the two breakpoints identified by CDSS are both closely aligned with the ground truth (500 and 1000), suggesting that CDSS is capable of maintaining outstanding performance in segmenting unstable multivariable time series.

\subsection{Penicillin Fed-batch Fermentation Process} 
\begin{table}[!t]
	\caption{The Process Variables of Penicillin Fed-batch Fermentation Process}  
	\centering
	\label{tab:penicillin variables}
	\begin{tabular}{l|l}
		\hline
		Variables&Descriptions \\
		\hline
		X1&Aeration rate \\
		\hline
		X2&Agitator power \\
		\hline
		X3&Substrate feed rate \\
		\hline
		X4&Substrate feed temperature \\
		\hline
		X5&Substrate concentrate \\
		\hline
		X6&Dissolved oxygen concentration \\
		\hline
		X7&Biomass concentration \\
		\hline
		X8&Penicillin concentration \\
		\hline
		X9&Culture medium volume \\
		\hline
		X10&$CO_2$ concentration \\
		\hline
		X11&PH  \\
		\hline
		X12&Reactor temperature  \\
		\hline
		X13&Generated heat \\
		\hline
		X14&Base flow rate \\
		\hline
		X15&Cold water flow rate \\
		\hline
	\end{tabular}
\end{table}

In this section, the penicillin fed-batch fermentation process is used to validate the efficiency of the CDSS model and the TC-GCN model. The experiment is implemented on PenSim v2.0 platform \cite{yuan2014soft}. The total simulation time is 400 hours. The sample interval is 0.025 h. The collected training and testing data sets each consist of 16000 samples. The two data sets are collected in two simulations in normal state, with different setting values of the initial substrate concentrate (15.0 g/L and 14.5 g/L respectively). According to the process knowledge, 0$\sim$45 h is the stage of biomass accumulation, and 45$\sim$400 h belongs to the continuous penicillin production stage. The variable descriptions of this process are listed in the Table \ref{tab:penicillin variables}. The dissolved oxygen concentration X6 is taken as the quality variable, denoted as Y. 
\begin{figure}[t]
	\centering
	\includegraphics[width=\columnwidth]{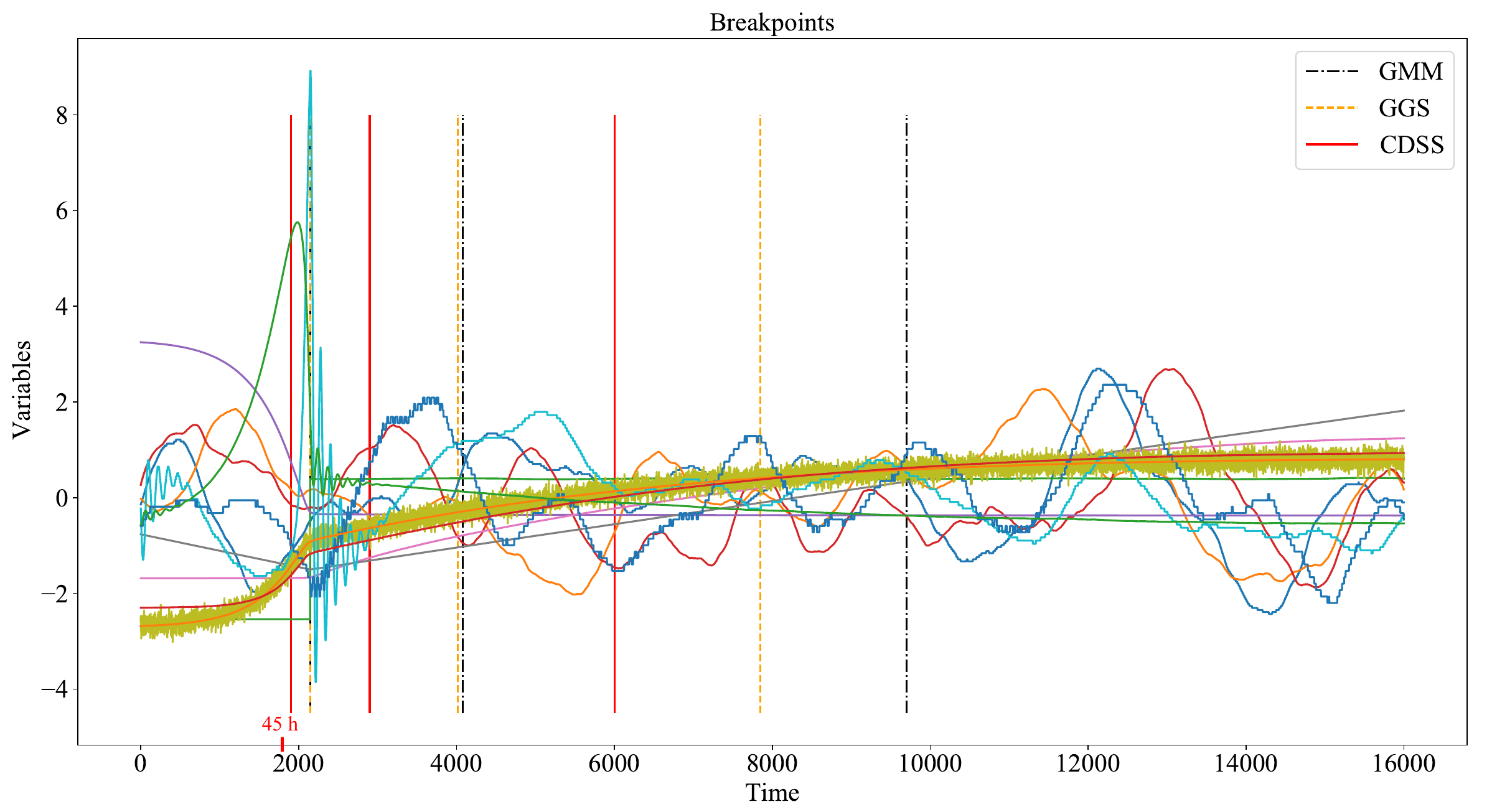}  
	\caption{The segmentation results of different models.} 
	\label{fig:breakpoints}   
\end{figure}

In the segmentation process, the initial window length $h$ is set to 1000. The maximum time lag $K$ is set to 3. The number of the convolutional kernels $m$ is set to 160. The coefficient $\zeta$ is 10. The two scale hyperparameters $\alpha$ and $\beta$ are respectively 2 and 1.2. As shown in Fig. \ref{fig:breakpoints}, the CDSS model segments the continuous 16000 testing samples into four phases: the obtained breakpoints are 1900, 2900, and 6000; the corresponding time stamps are respectively 47.5 h, 72.5 h, and 150 h. Fig. \ref{fig:breakpoints} also displays the segmentation results of the other two models, GMM and GGS. It can be seen that the first breakpoints found by the three methods are respectively 47.5 h (CDSS), 53.6 h (GMM), 53.6 h (GGS), and the corresponding true time stamp of phase transition is 45 h. Furthermore, the transitional stage between the breakpoint 1900 and 2900 is clearly recognized by CDSS (red solid lines). The results indicate that the breakpoints discovered by CDSS are more consistent with the ground truth, while the breakpoints found by GMM and GGS are more susceptible to the mean value.

After the segmentation, the TC-GCN models are established respectively using all the training data and the data in each divided phase. The numbers of hidden neurons in the two GC layers are 512 and 256 respectively. The numbers of hidden neurons in the MLP are 512 and 128 respectively. The batch size is 128. The learning rate is 0.001. The number of epochs is set to 2000. First, the input time series is expanded from 14 dimensional to 56 dimensional. Then, the fused features are obtained from the GC layer. Finally, the fused features and the residual information are processed through the MLP layer to obtain the predicted values.

\begin{figure}[t]
	\centering
	\includegraphics[width=\columnwidth]{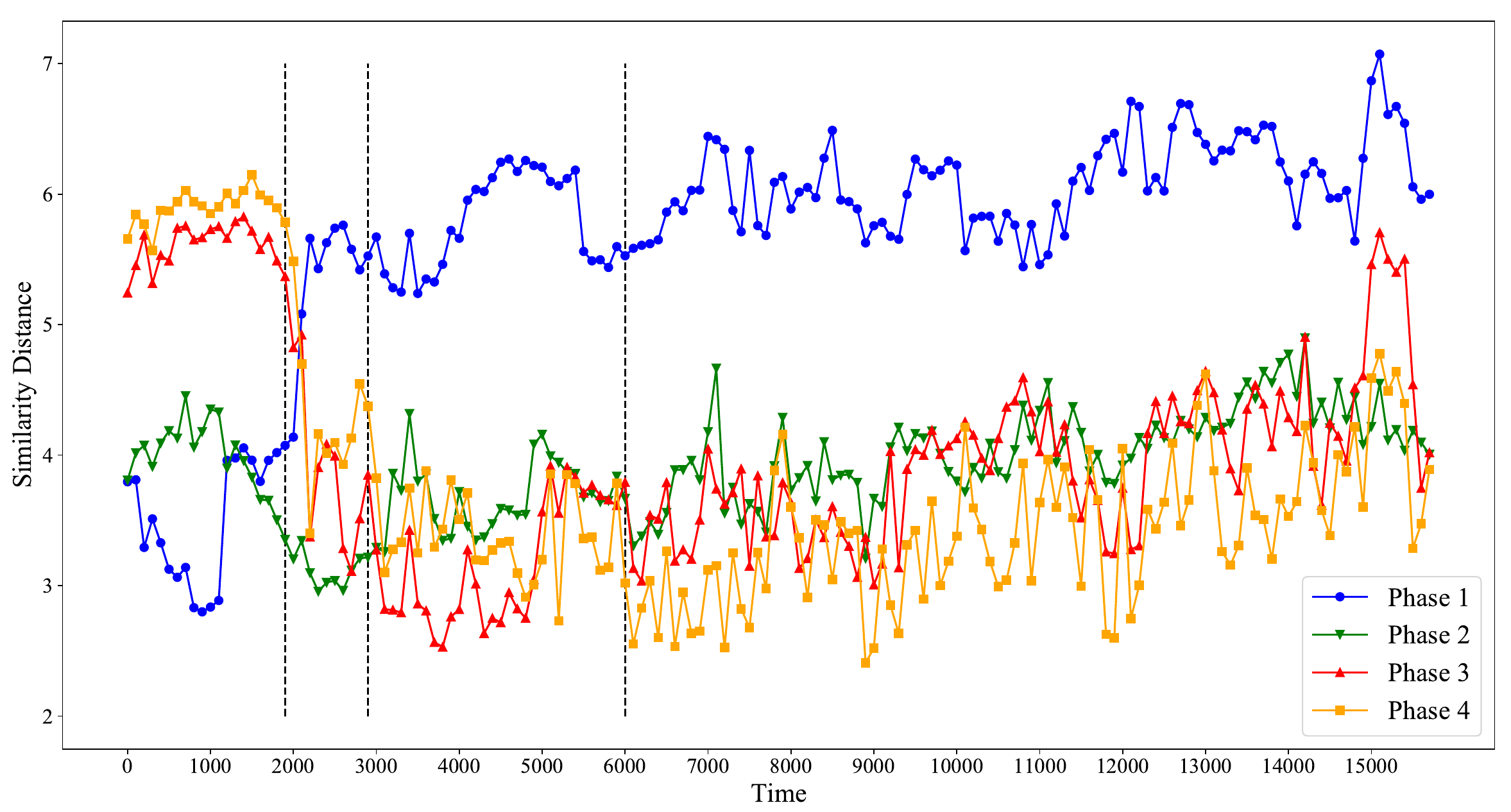}  
	\caption{The similarity distances between the testing samples and the samples in the segmented four phases.} 
	\label{fig:sd}   
\end{figure}

To complete quality forecasting, it is imperative to pair the test samples with their most appropriate phase, followed by the application of the corresponding predictive model to estimate the quality variable X6. Therefore, the similarity distances between the testing samples and the four phases are calculated by the CDSS model, which are displayed in Fig. \ref{fig:sd}. From Fig. \ref{fig:sd}, it can be clearly seen that over time, the phase with the minimum similarity distance to the testing samples varies. Moreover, the moments when the phase with the minimum similarity distance changes, is close to the breakpoints (black dashed lines) found by the CDSS model using the training data, indicating that the segmentation results have a certain degree of robustness.

\begin{table}[!t]
	\caption{The RMSEs of Different Models Using Various Training Data and Soft Sensing Models}  
	\centering
	\label{tab:soft sensing results} 
	\resizebox{0.5\textwidth}{!}{
		\begin{tabular}{c  c  c  c  c  c  c}
			\toprule
			\multirow{1}{*}{Phase}& \multicolumn{3}{c}{Using all the training data}&\multicolumn{3}{c}{Using the data of each phase}\\
			\cmidrule(lr){2-4} \cmidrule(lr){5-7} 
			& MLP & TC-GCN & TC-GCN(W) & MLP & TC-GCN & TC-GCN(W) \\
			\midrule
			Phase 1 & 0.174 & 0.148 & 0.168 & 0.089 & 0.070 & \textbf{0.062} \\
			Phase 2 & 0.219 & 0.266 & 0.306 & 0.233 & 0.202 & \textbf{0.196} \\
			Phase 3 & 0.151 & 0.185 & 0.164 & 0.118 & 0.089 & \textbf{0.082} \\
			Phase 4 & 0.143 & 0.127 & 0.105 & 0.034 & 0.027 & \textbf{0.027} \\
			Average & 0.153 & 0.149 & 0.136 & 0.069 & 0.055 & \textbf{0.052} \\
			\bottomrule
		\end{tabular}
		}  
\end{table}

During online testing, the testing samples are first extended, and then the phases to which the samples belong are evaluated based on the similarity distances. Then, the soft sensing model corresponding to the phase with the minimum similarity distance is used to predict the quality variables. Table \ref{tab:soft sensing results} shows the prediction results, including the root mean square errors (RMSEs) of predictions using different training data and various soft sensing models. Table \ref{tab:soft sensing results} shows the prediction results of three soft sensing models, MLP, TC-GCN, and TC-GCN (W). Among them, the difference between the TC-GCN and TC-GCN(W) is that the former uses the bool adjacency matrices of TCGs, while the latter uses the weighted adjacency matrices of TCGs.

\begin{figure}[t]
	\centering 
	\subfigure[MLP.]{  
		\includegraphics[width=0.9\columnwidth]{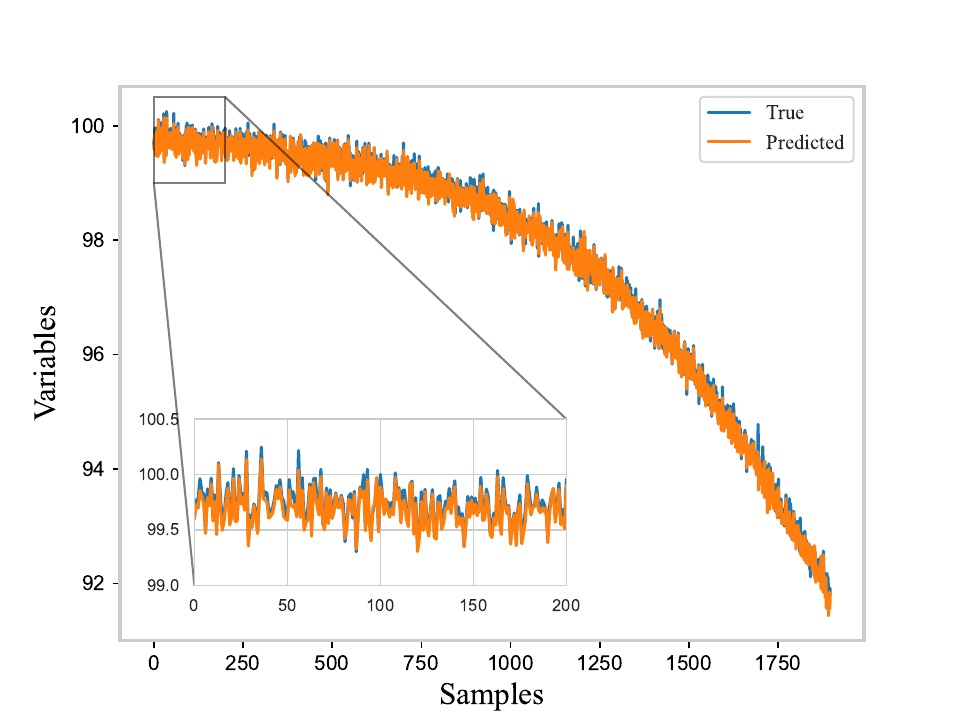} \label{fig:s_mlp_1} }
	\subfigure[TC-GCN(W).]{  
		\includegraphics[width=0.9\columnwidth]{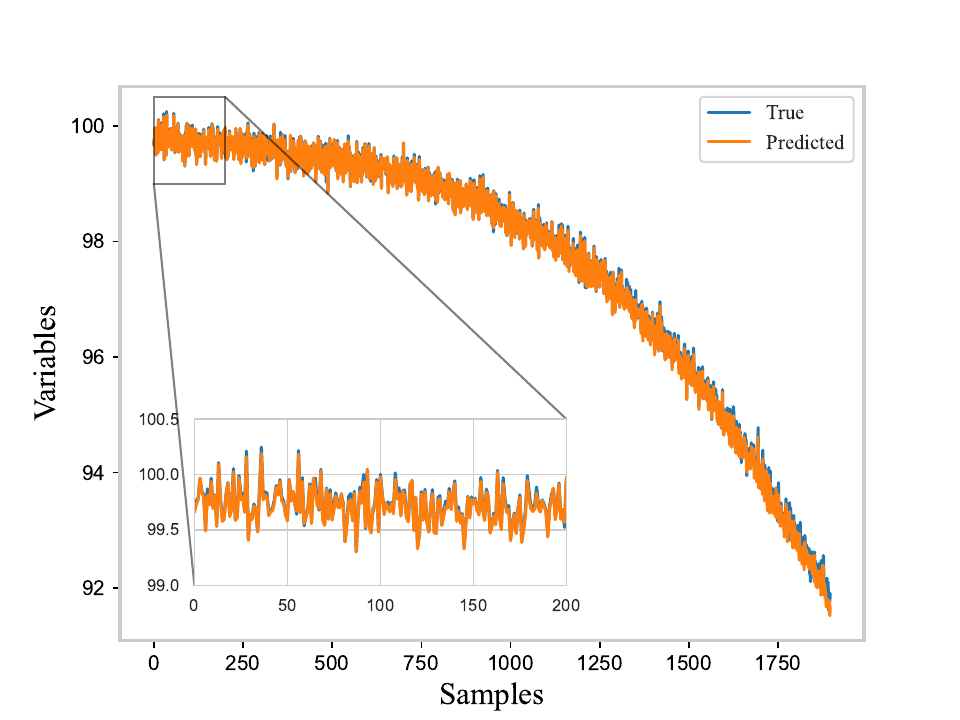} \label{fig:s_gcn(w)_1}}
	
	\caption{The predictions of X6 for the testing data in Phase 1.} \label{fig:predictions_phase1}
\end{figure}
 
\begin{figure}[t]
	\centering 
	\subfigure[MLP.]{  
		\includegraphics[width=0.9\columnwidth]{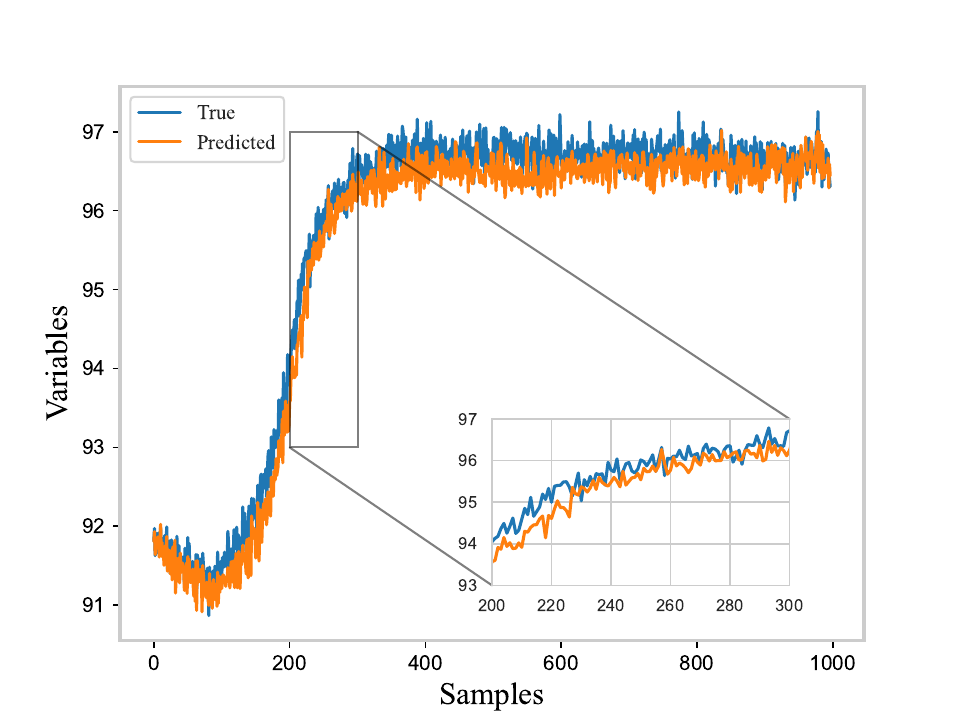} \label{fig:s_mlp_2}}
	\subfigure[TC-GCN(W).]{  
		\includegraphics[width=0.9\columnwidth]{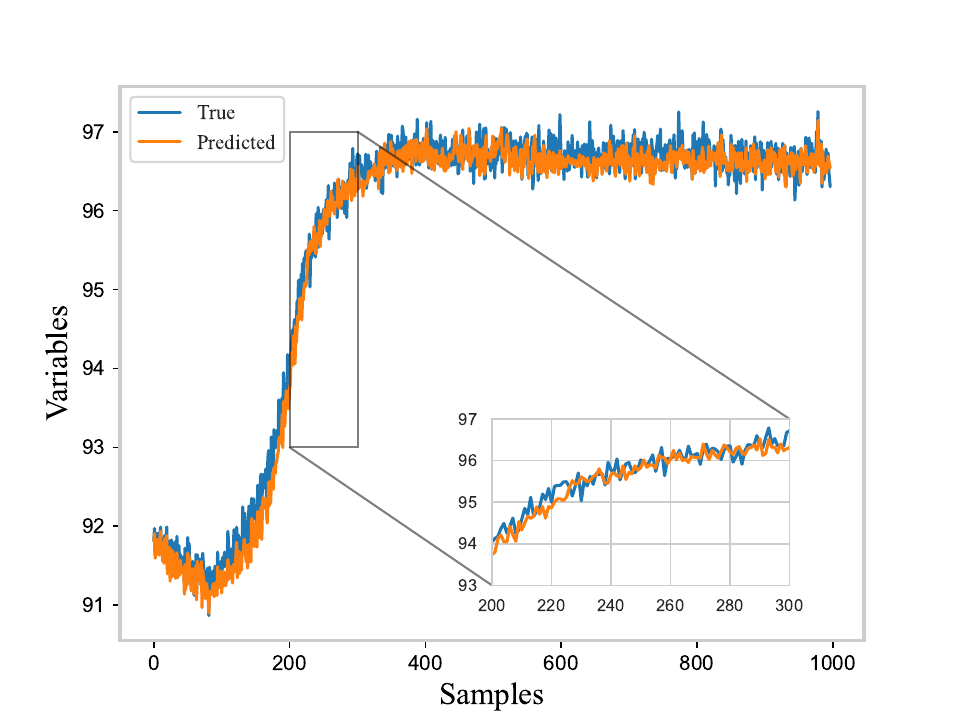} \label{fig:s_gcn(w)_2}}
	\caption{The predictions of X6 for the testing data in Phase 2.} \label{fig:predictions_phase2}
\end{figure}

By comparing the first three columns and the last three columns of Table \ref{tab:soft sensing results}, 
it can be evidently seen that employing phase segmentation for soft sensing enhances the predictive accuracy. As the results show, the soft models which are trained using the training data from one certain phase, are more capable in predicting the data in this phase. The improvements of phase 1, 3, and 4 are more significant. For the testing samples related to Phase 2, the TC-GCN and TC-GCN(W) which are trained with the training data in Phase 2, are both able to achieve better predictive performance. However, the MLP models cannot predict the quality variable more accurately, even though the segmenting-matching-predicting procedure is followed.

Comparing the prediction results in the last three columns, it can be seen that the TC-GCNs and the TC-GCN(W)s reach lower RMSEs than the MLPs, on the condition that the models are trained using the training data of each phase. For visualization, Fig. \ref{fig:predictions_phase1} and Fig. \ref{fig:predictions_phase2} respectively show the predictions of the MLPs and the TC-GCN(W)s for the testing data in Phase 1 and Phase 2. It indicates that the temporal-causal graph convolution improves the prediction performance for testing samples at all phases, and the improvement is more significant when using the weighted temporal-causal graphs. Nevertheless, when the models are trained using all the sequence data, the temporal-causal graph convolution may not guarantee the enhancement of the predictive performance. As Table \ref{tab:soft sensing results} displays, the addition of temporal-causal graph convolution enhances the predictive performance of the models for steady-state phases (like Phase 1, Phase 4). However, the performance of the predictive models for transitional phases tend to decline (like Phase 2, Phase 3).

\section{Conclusions}\label{sec:conclusions}
In this article, a new phase segmentation model called causality-driven sequence segmentation (CDSS) is proposed, and then a temporal-causal graph convolutional network (TC-GCN) is designed for soft sensing modeling. The CDSS model first discovers the local dynamic characteristics of causal relationships between variables and detects the breakpoints by identifying the abrupt changes of the causal mechanisms during phase transitions. The similarity distance metric, comprising causal similarity distance and stable similarity distance, is designed to quantify the divergence in causal mechanisms and stable conditions between samples from two distinct phases. For each phase, a TC-GCN model is established using the adjacency matrix of the temporal causal graph. The verification experiments are conducted on numerical examples and the penicillin fermentation industrial process. The segmentation results of the numerical examples demonstrate that CDSS is adept at identifying the phase transitions in both stationary and non-stationary multivariable series. As for the penicillin fermentation process, the breakpoints identified by CDSS are more accurately aligned with the ground truth when compared to other methodologies. Furthermore, the soft sensing model leveraging TC-GCN has exhibited a notable enhancement in predictive accuracy.

\ifCLASSOPTIONcaptionsoff
  \newpage
\fi

\bibliographystyle{IEEEtran}
\nocite{*}
\bibliography{SC_ref}

\end{document}